\def\year{2020}\relax
\documentclass[letterpaper]{article} %
\usepackage{aaai20}  %
\usepackage{times}  %
\usepackage{helvet} %
\usepackage{courier}  %
\usepackage[hyphens]{url}  %
\usepackage{graphicx} %
\usepackage{graphicx}  %
\usepackage{hyperref}
\nocopyright
\usepackage{natbib}
\usepackage{soul}
\usepackage{amsmath}
\usepackage{amssymb}
\usepackage{bm}
\usepackage{color}
\usepackage{listings}
\usepackage{multicol}
\usepackage{subcaption}
\usepackage{booktabs}
\usepackage{stfloats}

\usepackage[usenames,dvipsnames]{xcolor}

\lstdefinelanguage{Julia}%
  {morekeywords={abstract,break,case,catch,const,continue,do,else,elseif,%
      end,export,false,for,function,immutable,import,importall,if,in,%
      macro,module,otherwise,quote,return,switch,true,try,type,typealias,%
      using,while,@trace,@gen},%
   sensitive=true,%
   morecomment=[l]\#,%
   morecomment=[n]{\#=}{=\#},%
   morestring=[s]{"}{"},%
   morestring=[m]{'}{'},%
}[keywords,comments,strings]%

\usepackage{parskip}

\usepackage{mathtools}

\renewcommand{\vec}[1]{{\bm{#1}}}

\title{Sampling Prediction-Matching Examples in Neural Networks:\\A Probabilistic Programming Approach}
\author{
Serena Booth\thanks{Equal contribution}, Ankit Shah\footnotemark[1], Yilun Zhou\footnotemark[1], Julie Shah\\ 
Computer Science and Artificial Intelligence Laboratory\\ 
Massachusetts Institute of Technology\\
\{serenabooth, ajshah, yilun,  julie\_a\_shah\}@csail.mit.edu\\ 
}

\begin{document}

\maketitle
\begin{abstract}
Though neural network models demonstrate impressive performance, we do not understand exactly how these black-box models make individual predictions. This drawback has led to substantial research devoted to understand these models in areas such as robustness, interpretability, and generalization ability. In this paper, we consider the problem of exploring the prediction \emph{level sets} of a classifier using probabilistic programming. We define a prediction level set to be the set of examples for which the predictor has the same specified prediction confidence with respect to some arbitrary data distribution. Notably, our sampling-based method does not require the classifier to be differentiable, making it compatible with arbitrary classifiers. As a specific instantiation, if we take the classifier to be a neural network and the data distribution to be that of the training data, we can obtain examples that will result in specified predictions by the neural network. We demonstrate this technique with experiments on a synthetic dataset and MNIST. Such level sets in classification may facilitate human understanding of classification behaviors.\footnote{Code: \url{https://github.com/serenabooth/level-set-inference}}
\end{abstract}

\section{Introduction}

Are we comfortable with neural networks---with their high-dimensional, uninterpretable intermediary representations---making high stakes decisions such as who is released from jail or who receives a mortgage? As neural networks gain traction in commercial products, enabling people to understand these models becomes essential. To this end, many research communities have analyzed neural network properties. Research on fairness of neural networks aims to ensure models do not discriminate against protected groups~\citep{corbett2017algorithmic}. Research on robustness analyzes whether a model will perform drastically differently if the input is only slightly different from test data~\citep{carlini2019evaluating}. Research on interpretability tries to understand how decisions are made~\citep{doshi2017towards}. Finally, research on transfer learning explores if a neural network trained on one data distribution can work well or be adapted to work well on another distribution at test time~\citep{torrey2010transfer}. 

To complement these research approaches, we propose a technique for analyzing sets of examples of a given prediction confidence to evaluate a network's global performance. Our technique empirically evaluates prediction \emph{level sets}, or sets of a given confidence, through applying probabilistic programming~\citep{Cusumano-Towner:2019:GGP:3314221.3314642}. We first create a generative model and train a neural network to classify generated images. We then sample examples of a given confidence by applying Metropolis Hastings to infer a latent representation of an image which meets the specified classification confidence. We demonstrate our technique on two domains: a procedurally-generated synthetic domain and MNIST digits produced with a VAE or a GAN. Our results show that our inference successfully samples a set of examples for the given classification confidence in each domain. We further demonstrate how our technique can sample ``ambiguous" images from network decision boundaries.

\section{Related Work}

\subsection{Adversarial Attacks on Neural Networks}

Many approaches demonstrate that neural network predictions are locally unstable; i.e. an input can be slightly perturbed to make the network produce vastly different predictions.
\citet{szegedy2013intriguing} first demonstrated that images of seemingly random noise can receive confident predictions and a correctly classified image can be slightly perturbed to be confidently predicted as an incorrect class. Subsequent work has focused on more effective attack methods \citep[e.g.][]{goodfellow2014explaining, nguyen2015deep, carlini2017towards, athalye2017synthesizing, li2018second, athalye2018obfuscated} and better defenses against them \citep[e.g.][]{madry2017towards, cohen2019certified, lecuyer2019certified}. 

Remarkably, all these adversarial attacks produce examples that are out of distribution, and it is hard to characterize the distributional properties of such adversarial examples. Our work instead aims to produce ``adversarial'' examples in some \em known \rm distribution implicitly specified through a generative process. Our method is able to generate examples with arbitrary predictions (e.g. equally confident predictions across classes), while most adversarial attacks generate examples with confident predictions but incorrect labels. If the data generation models the distribution of the whole dataset except for a class $l$, and we generate examples according to that distribution but also enforce confident prediction in class $l$, we can uncover in-distribution adversarial examples. 

\subsection{Density Estimation}
Estimating the data distribution is an important and challenging task. Traditionally, parametric (e.g. Gaussian mixture model) and non-parametric (e.g. kernel density estimation \citep{rosenblatt1956remarks}) methods work well on low dimensional data, in which different features represent distinctively different meanings (such as age, salary, height, etc.). However, recent deep learning applications work on very high-dimensional datasets in which features are also highly correlated with each other. For example, an image can easily have tens of thousands of features (i.e. pixels), and nearby pixels are correlated. It is thus much more difficult, if even possible, for these traditional models to capture the data distribution and generate realistic-looking data. 

A commonly used model for estimating such high dimensional data distributions is an autoencoder, in which two neural networks respectively encode (compress) and decode (reconstruct) the image into and from a low-dimensional latent vector. Variational autoencoders (VAEs) \citep{kingma2013auto} additionally regularize the latent vectors to conform to a unit Gaussian distribution, which reduces the task of sampling from the data distribution to reconstruction from unit Gaussian samples. However, due to the training objective of minimizing $l_2$ loss, the images reconstructed for vision domain problems through a VAE are typically less sharp than the original images. 

Compared to autoencoders, generative adversarial networks (GANs) \citep{goodfellow2014generative} employ a generator-discriminator architecture designed to specifically regularize the generated image to stay within the data distribution. Consequently, mode collapse, in which the generated data lacks the variety found in the original data, is a common problem, and several methods \citep[e.g.][]{srivastava2017veegan, lin2018pacgan} have been proposed to solve the problem. In our experiments on a synthetic dataset with known low-dimensional features, we use the groundtruth distribution. On the MNIST image dataset, we use both a variational autoencoder and a GAN to generate images from the respective learned data distributions. 

\subsection{Confidence in Neural Networks}

Our technique explores confidence level sets in neural networks. However, prior work has shown many modern neural network architectures result in overconfident networks, with many incorrect predictions having undue high confidence~\citep{guo2017calibration}. To address the problem of overconfidence in neural networks, %
\citet{lee2017training} used a GAN to generate out-of-distribution data and regularize the classifier to have uniform confidence (i.e. complete ambivalence) on these examples. 
Several Bayesian formulations have also been formulated with different approximate inference methods \citep[e.g.][]{blundell2015weight, graves2011practical, balan2015bayesian}. 
\citet{gal2016dropout} drew a connection between the dropout layer \citep{srivastava2014dropout} and Bayesian formulation and used this to calculate uncertainty. 
\citet{lakshminarayanan2017simple} proposed an ensemble approach to capture model uncertainty. 

Our work is complementary to these models in that we do not attempt to alter the confidence of a trained neural network model, but instead expose examples with particular confidence values. As a result, if the model is over-confident, our sampling approach may return few or even no valid examples, within some specified distribution. Moreover, since we do not require access to anything beyond a prediction confidence (e.g. gradient information), our model can be used to study all of these confidence-calibrated models. 

\subsection{Probabilistic Programming}
Probabilistic programming consists of the emerging set of tools in which programming languages enable the definition of models and automate parts of inference. With popularity of deep learning, several frameworks \citep[e.g.][]{bingham2018pyro, salvatier2016probabilistic, tran2016edward} also have deep learning integration, in which a neural network model is used to define and/or learn a distribution. In our work, we use the state-of-the-art language Gen~\citep{Cusumano-Towner:2019:GGP:3314221.3314642}. One prior application of probabilistic programming has parallels to our approach: \citet{fremont2018scenic} designed a constrained probabilistic programming language, \textsc{Scenic}, which generates distributions over scenes produced from a generative model. While our work is inspired by \textsc{Scenic}, their system was designed to explore clear classification \emph{failures} and assist in dataset augmentation, whereas we apply probabilistic programming to explore confidence level sets---including classification successes. 

\section{Methodology}
Given a classifier $f: X \rightarrow \Delta ^ L$ that maps a data point to the probability simplex of $L$ classes, the overall goal is to find an input $\vec x\in X$ such that $f(\vec x)=\vec p$ for some particular prediction confidence $\vec p \in \Delta ^ L$. A common approach to achieve this is to start with some initial guess $\vec x_0$, and iteratively optimize the function $\mathcal L(x)=d(f(\vec x), \vec p)$ using gradient-based method, for some distance metric $d(\cdot, \cdot)$. One such example distance metric is the total variation distance. This method is reminiscent of popular adversarial attacks. However, there is no guarantee that such the resulting $\vec x^*$ will still remain in the original data distribution. 

Thus, we introduce a data distribution $p(\vec x)$ and consider the inference problem of sampling from the posterior 
\begin{align*}
p(\vec x|f(\vec x)=\vec p)\propto p(\vec x)p(f(\vec x)=\vec p|\vec x). 
\end{align*}
Note that the likelihood is actually a delta function; this makes sampling infeasible because in general the set $\{\vec x: f(\vec x)=\vec p\}$ has measure 0. To solve this problem, we relax the formulation by introducing $\tilde{\vec p}_\vec x$, which is sampled from a Dirichlet distribution with parameters proportional to $f(\vec x)$: 
\begin{align*}
    \tilde{\vec p}_\vec x\sim \mathrm{Dir}(\alpha f(\vec x)), 
\end{align*}
where $\alpha$ is a hyper-parameter to be set to determine the tolerance of an inaccurate $\tilde{\vec p}_x$: the lower the value of this hyperparameter, the more tolerant the sampling procedure. 

In addition, rather than sampling directly from the data distribution, we sample in the latent factor space $Z$, which can be mapped to $X$ via a deterministic reconstruction function $g: Z \rightarrow X$. We use the notation $\tilde{\vec p}_\vec z \doteq \tilde{\vec p}_{g(\vec x)}$. We assume the latent space distribution $p(\vec z)$ is much easier to sample from; for example, if our generative model is a GAN or VAE, this distribution is a unit Gaussian. 

To summarize, given
\begin{align}
    \vec z &\sim p(\vec z),\\
    \vec x &= g(\vec z),\\
    \tilde{\vec p}_\vec z&\sim\mathrm{Dir}(\alpha f(\vec x)),\\
    p(\vec z\,|\,\tilde{\vec p}_\vec z=\vec p)&\propto p(\vec z)p(\tilde{\vec p}_\vec z=\vec p \,| \,\vec z), \label{posterior}
\end{align}
where $\alpha$ is the hyper-parameter and $\vec p$ is the target prediction, we wish to sample from posterior $p(\vec z\,|\,\tilde{\vec p}_\vec z=\vec p)$, and reconstruct $\vec x$ from $\vec z$ using $g(\cdot)$. We see that the posterior distribution is large only if the prior is large as well, limiting our sampled $\vec z$ (and therefore the reconstructed $\vec x$) to the generative distribution $p(\vec x)$. In addition, we have $\lim_{\alpha\rightarrow\infty} ||f(g(\vec z))-\vec p||=0$ with probability 1, meaning that prediction of the samples will get closer to the target prediction as we increase $\alpha$. 

To sample from the posterior, we use the Metropolis-Hastings algorithm. We start from $\vec z^{(0)}\sim p(\vec z)$ and then use a Gaussian proposal function centered on the current $\vec z$ with symmetric covariance matrix (i.e. $kI$ for $k>0$). To exploit parallelism of computer architectures, we start with $N$ particles, and we sample for $T$ time steps. In this process, we conservatively choose a relatively large $T$ so that we sample beyond the requisite ``burn-in" time. For each particle, the last sample $z^{(T)}$ of the trajectory is selected. Finally, we select $n\leq N$ such unweighted samples using resampling with replacement. 

\begin{figure*}[!htb]
    \centering
    \begin{subfigure}[b]{0.30\textwidth}
        \centering
        \includegraphics[width=\textwidth,trim={1cm 1cm 1cm 1cm},clip]{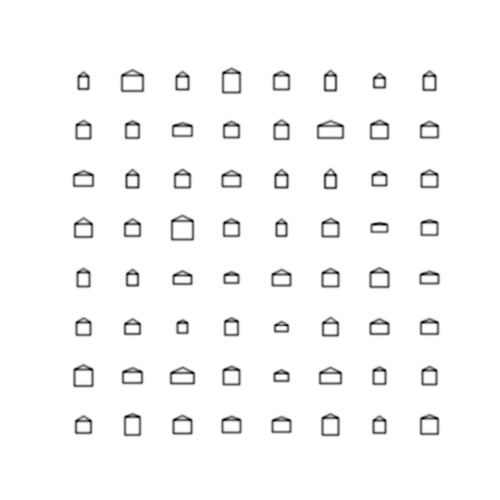}
        \caption{Predicted houses, $\beta=0.999$}
        \label{sample-houses}
    \end{subfigure}
    \hspace{5mm}
    \begin{subfigure}[b]{0.30\textwidth}
        \centering
        \includegraphics[width=\textwidth,trim={1cm 1cm 1cm 1cm},clip]{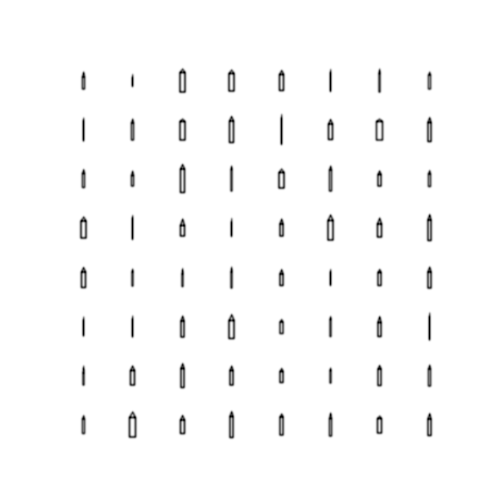}
        \caption{Predicted rockets, $\beta=0.001$}
        \label{sample-rockets}
    \end{subfigure}
    \hspace{5mm}
    \begin{subfigure}[b]{0.30\textwidth}
        \centering
        \includegraphics[width=\textwidth,trim={1cm 1cm 1cm 1cm},clip]{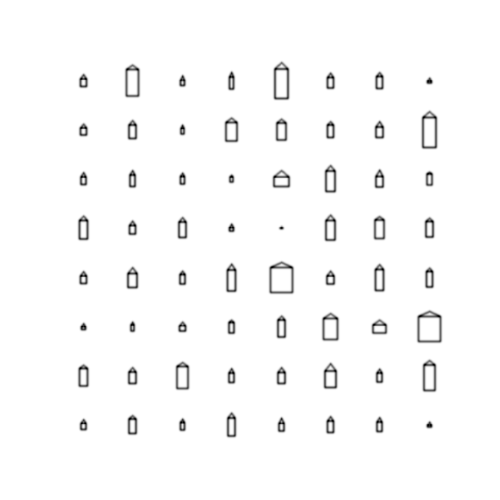}
        \caption{Ambiguous images, $\beta=0.5$}
        \label{sample-ambiguous}
    \end{subfigure}
    \caption{Given our trained classifier and generative world, we constrain $\beta$ to infer images of a given confidence level. Qualitatively, the generated ambiguous images appear to be primarily either extremes drawn from the distribution (e.g., small values for rectangle height $h$) or images with aspect ratios in between those of prototypical houses and rockets as shown in Figure \ref{fig:prototypical-house-rocket}.}
    \label{fig:confident-vs-ambiguous-houserocketdomain}
\end{figure*}

\section{Experiments}

We conduct experiments on two datasets: a synthetic image dataset in which images look either like a ``house'' or a ``rocket'' (see figure \ref{fig:prototypical-house-rocket}), and the MNIST dataset. For both datasets, we either have access to the true data distribution or learn a data distribution $p(\vec z)$ over the latent space. We then sample $n$ latent vectors $\{\vec z_i\}_{i=1}^n$ from the posterior distribution (Eqn \ref{posterior}) with certain known or learned data distribution (i.e. prior) and specified target prediction $\vec p$, and reconstruct $\{\vec x_i\}_{i=1}^n$ using the reconstruction function $g(\cdot)$. 

To evaluate the quality of our sampled images, we compute the predicted probabilities: $\{\tilde{\vec p}_i=f(\vec x_i)\}_{i=1}^n$. We then calculate average deviation from target over class according to
\begin{align}
\Delta\doteq\frac{1}{n}\frac{1}{|L|}\sum_{i=1}^n\sum_{l\in L}|\vec p_l-\tilde{\vec p}_{i, l}|, \label{delta}
\end{align}
where $\tilde{\vec p}_{i, l}$ is the predicted probability of $\vec x_i$ for class $l$. 

\subsection{Rockets or Houses: A Synthetic Geometric World}

\begin{figure}[!b]
    \centering
    \includegraphics[width=0.29\columnwidth]{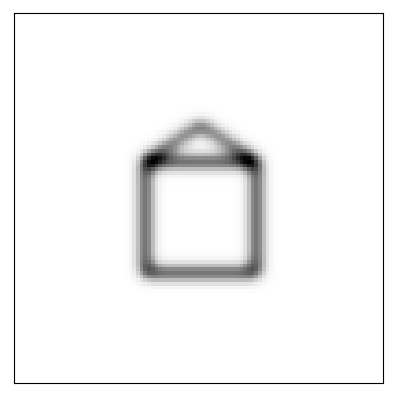}
    \includegraphics[width=0.29\columnwidth]{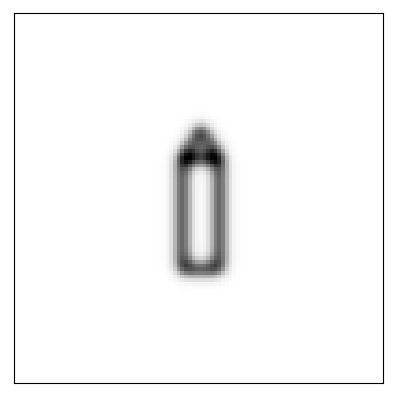}
    \includegraphics[width=0.29\columnwidth]{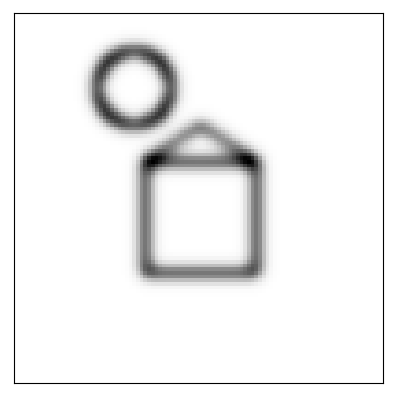}
    \caption{Left: the prototypical house image; middle: the prototypical rocket image; right: introduction of out of distribution circle geometry. Images have a Gaussian blur applied to them for ease of classification.}
    \label{fig:prototypical-house-rocket}
\end{figure}

We construct a synthetic world where simple rules govern how an image of a triangle on top of a rectangle is drawn. Some of the images look like ``houses'' when  a short triangle is put on top of a wide rectangle. Others look like ``rockets'' when a tall triangle is put on top of a slim rectangle. See figure \ref{fig:prototypical-house-rocket} for prototypical images. Each image is rendered from three parameters: rectangle width $w$, rectangle height $h$, and triangle height $t$, which are generated from a mixture model conditioned on a latent variable $c$ representing whether the image should be a house or a rocket. 
Specifically, to draw an image we follow the generative process: 
\begin{align*}
    c &\sim \mathrm{Ber}(0.5),\\
    w|(c=0) &\sim \mathrm{No}_+(10, 5^2),\\
    h|(c=0) &\sim \mathrm{No}_+(30, 10^2),\\
    t|(c=0) &\sim \mathrm{No}_+(8, 2^2),\\
    w|(c=1) &\sim \mathrm{No}_+(30, 10^2),\\
    h|(c=1) &\sim \mathrm{No}_+(30, 10^2),\\
    t|(c=1) &\sim \mathrm{No}_+(10, 2^2),\\
    \vec z & \doteq [w, h, t]
\end{align*}
where $\mathrm{No}_+(\mu, \sigma^2)$ is the normal distribution truncated to the positive support. After rendering these images, we subsequently applied a Gaussian blur for for easier classification.

\begin{figure}[!htb]
    \centering
    \includegraphics[width=\columnwidth,trim={7cm 2cm 2cm 1cm},clip]{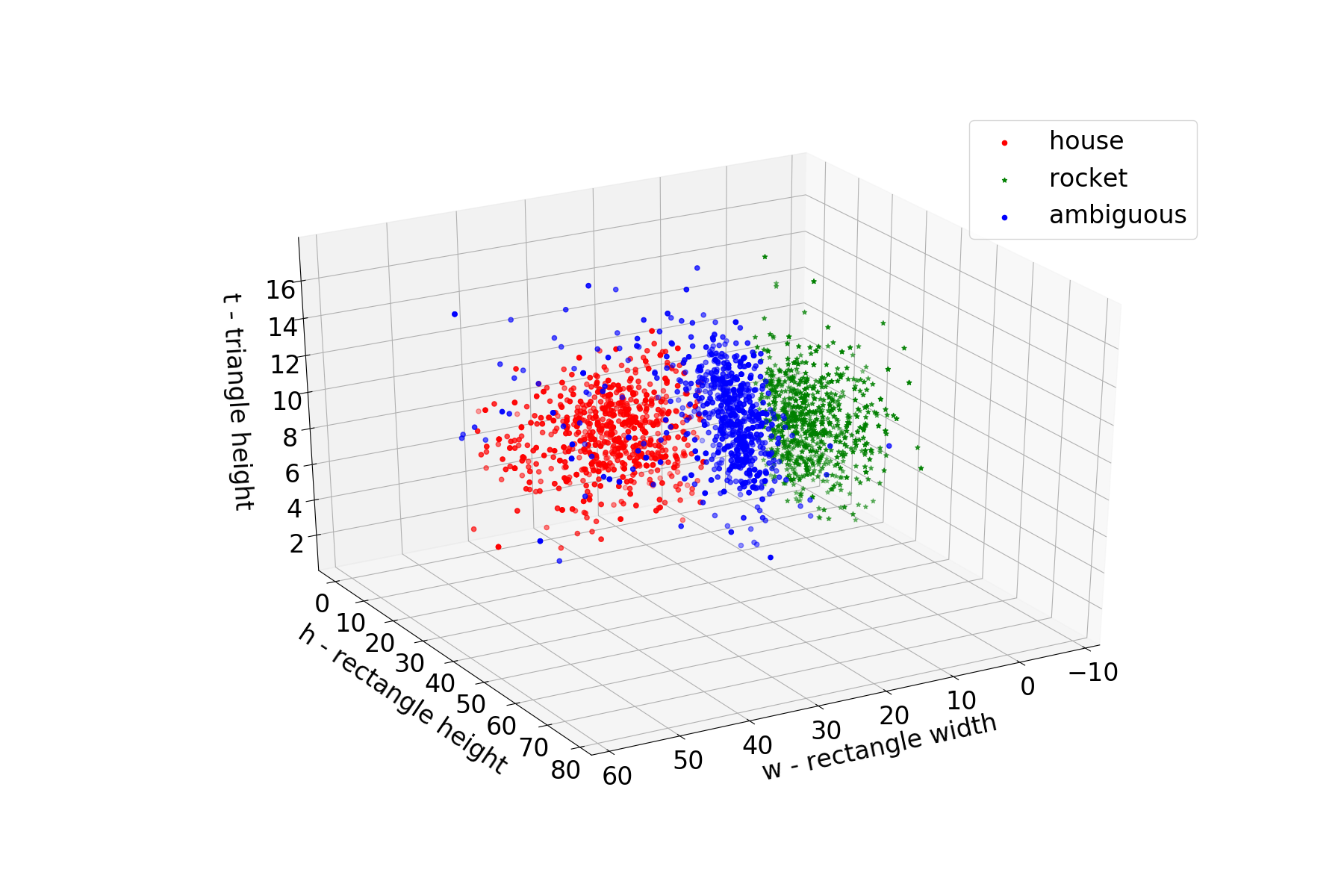}
    \includegraphics[width=\columnwidth,trim={7cm 2cm 2cm 1cm},clip]{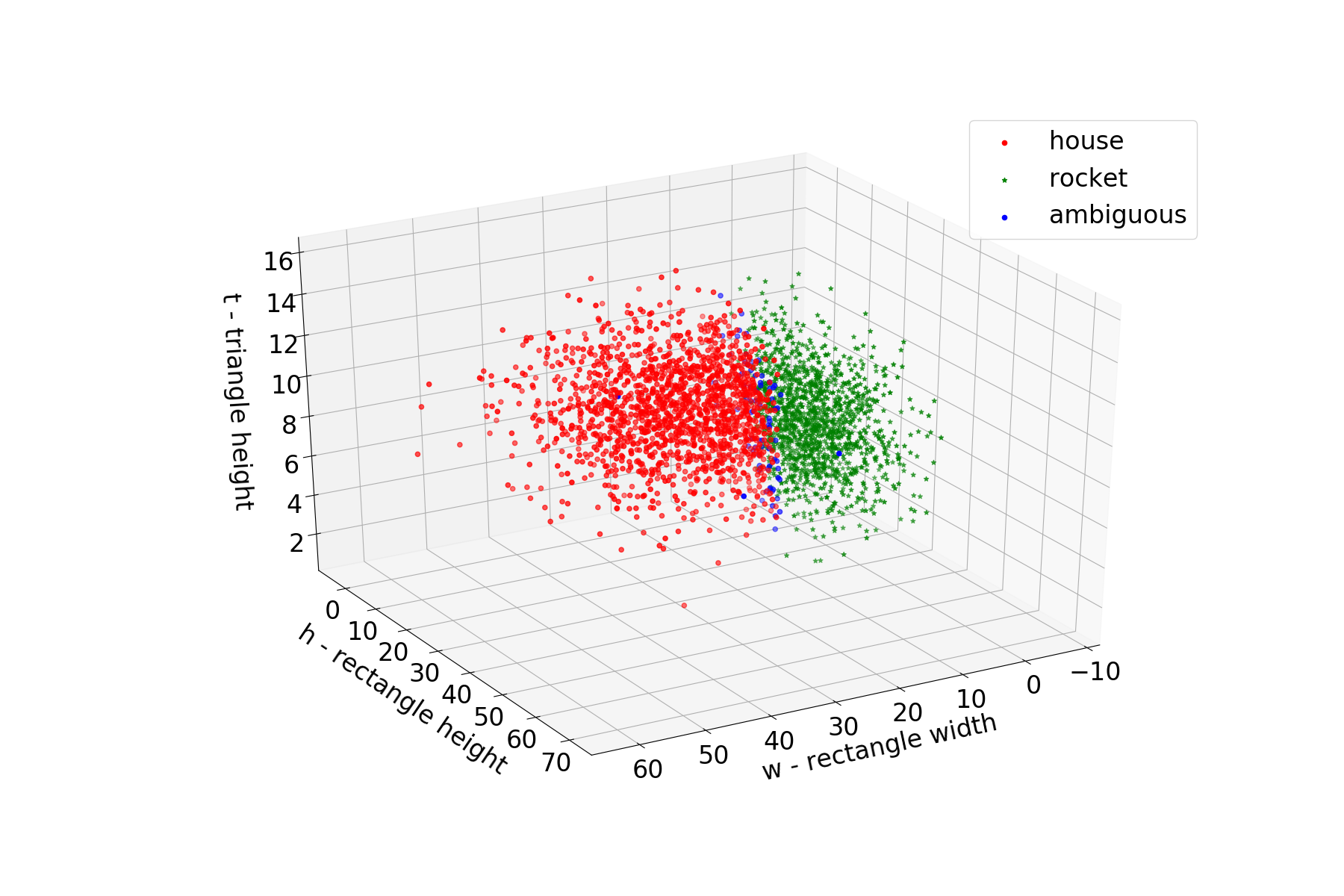}

    \caption{Above: three-dimensional latents discovered through sampling the posterior as predicted confidently houses, confidently rockets, or ambiguous. Below: 3000 latents generated directly by sampling the prior distributions. Ambiguous images have confidences ranging between 45\% and 55\% for each class. Indeed, we discover many more ambiguous examples through our targeted inference approach.}
    \label{fig:latents-house-rocket}
\end{figure}

Using the synthetic images and the known groundtruth labels ($c$), we train a CNN-based classifier on 16,000 such images. Our classifier achieves 97.3\% accuracy (94.7\% precision, 97.4\% recall). We then use our method to sample examples with target prediction $\vec p=[0.5, 0.5]$, with $\alpha=10$. Note that in this binary case, the Dirichlet distribution reduces to the Beta distribution. 

Figure \ref{fig:confident-vs-ambiguous-houserocketdomain} plots some of the images sampled from the posterior distribution with $\beta$ constrained to produce images which confidently classify as houses ($\beta = 0.999$), images which confidently classify as rockets ($\beta = 0.001$), and images which classify as ambiguous ($\beta = 0.5$). Empirically, these images indeed appear to be houses, rockets, or ambiguous images. For generated houses, the mean classifier confidence was 94.9\% (minimum 84.1\%); for generated rockets, the mean classifier confidence was 94.9\% (minimum 86.4\%). The average generated image resulted in a classification confidence of 51.0\%. This result is also confirmed in Figure \ref{fig:latents-house-rocket}, which visualizes the parameters of these ambiguous images rendered among the mixtures of two Gaussians. We compare our inference approach for selecting images from the three level sets to a set of randomly generated latents. We find many more ambiguous images through our inference approach than when directly sampling latents from the respective distributions.

Our method is not limited to test data from the distribution. We also demonstrate its potential for understanding behaviors of a trained classifier on different data distribution. To this end, we fix the classifier, trained on these ``house'' or ``rocket'' images, but we introduce a circle to the geometry rendering. A circle is randomly overlaid on an image as shown in Figure \ref{fig:prototypical-house-rocket}. We introduce the parameters circle radius $r$ and circle center $(x,y)$ into our generative process:

\begin{align*}
    r &\sim \mathrm{No}_+(20, 10^2),\\
    x &\sim \mathrm{U}(20, 40)\\
    y &\sim \mathrm{U}(20, 40),\\
    \vec z & \doteq [w, h, t, r, x,y]
\end{align*}
where $w, h,$ and $t$ are defined as before, and $\mathrm{U}(a, b)$ is a uniform distribution.

Using this new generative function, we again infer images from our three classes: those for which the classifier confidently predicts houses, those for which the classifier confidently predicts rockets, and those for which the classifier has low prediction confidence. We then evaluate how the presence of this out-of-distribution geometry affects classification predictions, and we find it has minor effects. For generated rocket images, average confidence in the rocket prediction falls 1.6\% when circles are rendered; accuracy is unchanged. For generated house images, average confidence in the house prediction increases 1.1\%; accuracy is likewise unchanged. For generated ambiguous images, average confidence in predictions increases 4.9\% when circles are rendered. Curiously, when we infer ambiguous images with the circles rendered, the accuracy of the classifier is 15\% higher than when the circles are not rendered.

\subsection{MNIST}
On the MNIST dataset, we learn to approximate the data distribution using both a variational autoencoder (VAE) and a generative adversarial network (GAN) as the source data distribution is not available. Both models are trained with a 5-dimensional latent space. Hence, we have two induced data distribution $p_{VAE}(\vec x)$ and $p_{GAN}(\vec x)$, derived from $p_{VAE}(\vec z)=p_{GAN}(\vec z)=\mathrm{MVN}(\vec 0, I)\in \mathbb R^5$. For each VAE and GAN prior, we studied three target predictions: 
\begin{enumerate}
    \item ambiguous, with $\vec p_l=0.1 \,\,\,\forall l\in\{0, 1, ..., 9\}$;
    \item 1vs7, with $\vec p_1=\vec p_7=0.46$ and $\vec p_l=0.01 \,\,\, \forall l\neq 1, 7$;
    \item 8vs9, with $\vec p_8=\vec p_9=0.46$ and $\vec p_l=0.01 \,\,\, \forall l\neq 8, 9$.
\end{enumerate}

Table \ref{mnist_deviation} shows the average prediction deviation $\Delta$ as defined in Eqn \ref{delta}. We can see that in general these models are indeed sampling data points around the specified target prediction. While they seem to have comparable performance, we note in the qualitative evaluations that sampling with a GAN prior will reduce much less valid samples than with a VAE prior. 

\subsubsection{VAE Visualization}

\begin{figure*}[!htb]
    \centering
    \begin{subfigure}[b]{0.3\textwidth}
        \centering
        \includegraphics[width=\textwidth]{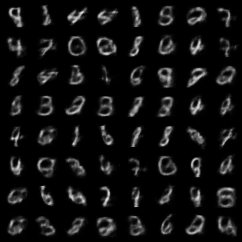}
    \end{subfigure}
    \hspace{5mm}
    \begin{subfigure}[b]{0.3\textwidth}
        \centering
        \includegraphics[width=\textwidth]{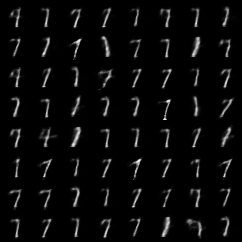}
    \end{subfigure}
    \hspace{5mm}
    \begin{subfigure}[b]{0.3\textwidth}
        \centering
        \includegraphics[width=\textwidth]{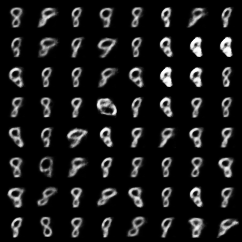}
    \end{subfigure}
    \begin{subfigure}[b]{0.3\textwidth}
        \centering
        \includegraphics[width=\textwidth,trim={0.33cm 0cm 0.33cm 0cm},clip]{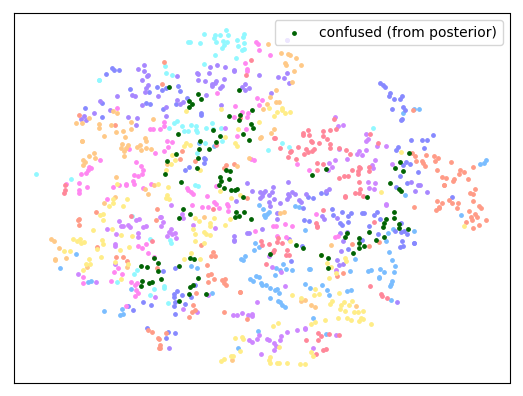}
        \caption{Ambiguous target}
        \label{vae-amblatent}
    \end{subfigure}
    \hspace{5mm}
    \begin{subfigure}[b]{0.3\textwidth}
        \centering
        \includegraphics[width=\textwidth,trim={0.33cm 0cm 0.33cm 0cm},clip]{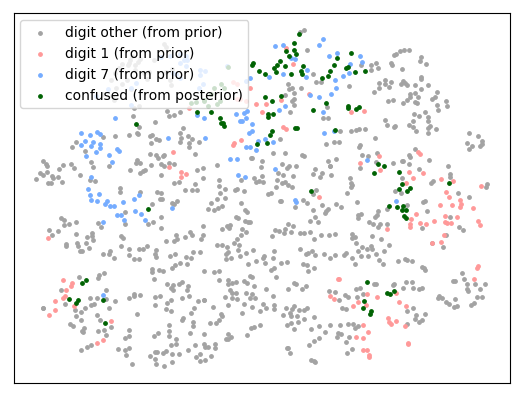}
        \caption{1vs7 target}
        \label{vae-1v7latent}
    \end{subfigure}
    \hspace{5mm}
    \begin{subfigure}[b]{0.3\textwidth}
        \centering
        \includegraphics[width=\textwidth,trim={0.33cm 0cm 0.33cm 0cm},clip]{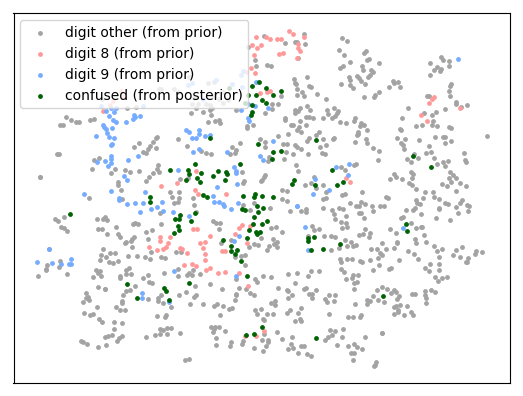}
        \caption{8vs9 target}
        \label{vae-8v9latent}
    \end{subfigure}
    \caption{Images inferred for specified prediction confidences sampled from the VAE's latent space, and visualizations of their presence in the latent space. Dark green dots represent selected ambigious samples. Left: images showing equal confidence in predictions for all digits $0,1,\dots,9$. Middle: images showing equal confidence for 1 and 7 classifications. Right: images showing equal confidence for 8 and 9 classifications. }
    \label{vae-image}
\end{figure*}

\begin{table}[t]
    \centering
    \begin{tabular}{c|c|c}\toprule
         & VAE & GAN\\\midrule
        ambiguous & 1.83\% & 5.93\%\\
        1vs7 & 6.22\% & 7.92\%\\
        8vs9 & 7.66\% & 5.31\%\\\bottomrule
    \end{tabular}
    \caption{Average deviation for different generative model and different prediction targets. }
    \label{mnist_deviation}
\end{table}

For qualitative evaluation, we visualize both the generated images and the latent space (reduced to 2 dimensions using t-SNE \citep{maaten2008visualizing}). 

The images sampled from the VAE prior for each class are shown in Figure \ref{vae-image}. For the ambiguous target, we see that the depicted images are often the result of several digits overlaid on each other and lacking in intensity. For the 1vs7 target, we see that most samples are indeed ``between'' digit 1 and digit 7, in that the horizontal stroke is short but visible in most cases. For the 8vs9 target, the lower circle is in general smaller than the upper circle, which is also visually between digit 8 (which should have equally-sized circles) and digit 9 (in which the lower circle degenerates to a vertical stroke). 

With the same VAE model, the sampled latent vectors for the three targets are plotted in Figure \ref{vae-image} after reduction to 2 dimensions using t-SNE. For Figure \ref{vae-amblatent}, the lightly colored dots represent latent vectors sampled from the prior distribution (i.e. 5D unit Gaussian), color-coded according to the classifier prediction, and the dark green dots represent latent vectors sampled from the posterior (Eqn \ref{posterior}). For Figure \ref{vae-1v7latent} and \ref{vae-8v9latent}, the gray, blue and red dots represent latent vectors sampled from the prior distribution, and are color-coded according to classifier predictions. The dark green dots represent latent vectors sampled from the posterior. 

We can see that the sampled latent vectors for the ambiguous case indeed lie at the intersections of several colored regions, though probably not between \em all \rm regions, because it is hard for an image to look like all the 10 digits at the same time. In addition, they are also quite centrally located, indicating a high likelihood of drawing this digit according to VAE. For the latent vectors of 1vs7 and 8vs9, they are generally ``in-distribution'', although they do tend to occupy a smaller probability mass. This smaller mass is expected, as most of the highly likely images (including those seen in the training set) are indeed unambiguously a single digit. However, the sampled latent vectors are also not completely out of distribution, which is commonly believed to be the case for adversarial examples \citep{li2019nattack}. 

\subsubsection{GAN Visualization}
\begin{figure}[!htb]
    \centering
    \begin{subfigure}[b]{0.8\columnwidth}
        \centering
        \includegraphics[width=\textwidth]{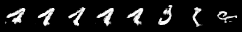}
        \caption{Ambiguous target}
        \label{gan_amb}
    \end{subfigure}
    \begin{subfigure}[b]{0.8\columnwidth}
        \centering
        \includegraphics[width=\textwidth]{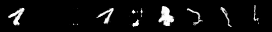}
        \caption{1vs7 target}
        \label{gan_1v7}
    \end{subfigure}
    \begin{subfigure}[b]{0.8\columnwidth}
        \centering
        \includegraphics[width=\textwidth]{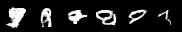}
        \caption{8vs9 target}
        \label{gan_8v9}
    \end{subfigure}
    \caption{Sampled images for three targets with GAN-learned data distribution. }
    \label{gan-image}
    \vspace{-5mm}
\end{figure}

For posterior sampling with a GAN as the learned generative model, the final resampling step (which is done with replacement) produces less than 10 distinct latent vector samples. Thus, we show the reconstructed images from all of them in Figure \ref{gan-image}. 

The small number of distinct samples in the GAN model results from the classifier being overly confident on most samples, an issue discussed in the Related Work section. This overconfidence makes it increasingly challenging to sample an image which produces ambiguous predictions from the generative model. 

We see that in comparison with the VAE, the images reconstructed with the GAN model are much sharper. The GAN generator explicitly penalizes the production of blurry images that do not look like real images. Therefore, we believe that the sampler is using a different strategy: rather than sampling blurred images (which would have high likelihood under VAE but low likelihood under GAN), it samples images that look like hand-drawn digits but are not actual digits (e.g. last image of Figure \ref{gan_amb} and second image of Figure \ref{gan_8v9}). One direct extension of this work is to try to use GAN model on some other handwritten datasets such as Omniglot \citep{lake2013one} and Kuzushiji-MNIST \citep{clanuwat2018deep}. 

\section{Conclusion and Future Work}
In this paper, we propose a method to sample data that have known classification prediction values. Compared to directly performing unconstrained gradient descent from an initial data point, as is the case in most adversarial attack work, our method also regularizes the resulting examples such that the sampled data has high likelihood under some specified data distribution. To achieve this, we formulate the inference problem as a Bayesian one, and we use probabilistic programming to sample from the posterior. In addition to selecting ``in-distribution'' samples, our method operates globally: the initial starting point does not affect the final result after sufficiently many burn-in samples have been drawn and discarded. On both a synthetic dataset and the MNIST dataset, we demonstrate that our method can indeed sample data points close to the specified target while remaining likely under the given distribution. In addition, our technique may be useful for identifying overconfident networks: the lack of diversity of ambiguous samples with the MNIST GAN distribution indicates that the MNIST classifier may be overly confident on GAN-generated images. 

There are several future directions for this work. First, MNIST is a relatively easy task with visually distinct classes. We will evaluate our technique on harder datasets such as CIFAR-10 or CIFAR-100 for which ambiguous labels (e.g. car vs truck for CIFAR-10 or different classes within one super-class in CIFAR-100) are more likely. Moreover, we can verify the effectiveness of confidence calibration \citep{guo2017calibration} of the classifier using our model. 

Our method could be also extended for in-distribution adversarial example discovery. For such discovery, we would require or learn a generative model for the entire dataset with examples for one class removed. Given this distribution, we would then sample examples from the posterior given the inference target of being highly confident as the missing class. 

Finally, we can use this method to create neural network models with notions of ambiguity similar to those of a human. Specifically, we plan to iteratively train a classifier and augment the dataset to incorporate labeled examples drawn from the ambiguous sets. In the end, the classifier will imitate human prediction both in certain and in ambiguous cases and will hopefully make only human-like mistakes, which could be valuable for use in sensitive domains.

\section{Acknowledgements}
We thank Alex Lew, Marco Cusumano-Towner, Christian Muise, and Hendrik Strobelt for fruitful discussions.

\bibliographystyle{aaai}
\bibliography{bibliography}

\end{document}